\documentclass[11pt]{article}

\usepackage{acl}
\usepackage{times}
\usepackage{latexsym}
\usepackage[T1]{fontenc}
\usepackage[utf8]{inputenc}
\usepackage{microtype}
\usepackage{graphicx}
\usepackage{booktabs}
\usepackage{array}

\usepackage[most]{tcolorbox}
\usepackage{cuted}
\usepackage{changepage}

\newtcolorbox{keymessage}{
  enhanced,
  breakable,
  width=\textwidth-15pt,
  colback=gray!3,
  colframe=purple,
  boxrule=0.8pt,
  leftrule=2.8pt,
  rightrule=2.8pt,
  toprule=0pt,
  bottomrule=0pt,
  arc=1mm,
  left=5pt,
  right=5pt,
  top=5pt,
  bottom=5pt,
}

\title{From Found to Designed:\\Concepts as a Design Axis for Large Language Models}

\author{
  Chen Shani \\
  Tel Aviv University \\
}

\begin{document}
\maketitle

\begin{strip}
\begin{adjustwidth}{15pt}{0pt}
\vspace{-80pt}
\begin{keymessage}
\centering
\textbf{Concepts are currently discovered. We argue they should be designed.}\\
LLMs encode concepts implicitly, through distributed associations. 
We call for architectures that treat concepts as explicit, compositional objects, and map the design space for building them.
\end{keymessage}
\vspace{-10pt}
\end{adjustwidth}
\end{strip}

\begin{abstract}
Large language models (LLMs) encode rich concept-like information, but represent it implicitly through distributed statistical associations rather than as explicit, structured, compositional concepts. Consequently, concept-level structure is typically \emph{found} rather than \emph{designed}: it is recovered after training through probing or dictionary learning, with no architectural guarantee of stability, compositionality, controllability, or alignment with human conceptual organization.
We organize concept-aware interventions along two dimensions: whether concept structure is internally induced or externally grounded, and the stage of the pipeline where it is introduced. This taxonomy reveals three broad patterns: inference-time approaches remain comparatively underexplored, related ideas have developed largely in isolation across pipeline stages, and externally grounded methods span the entire pipeline despite often being described under different terminology. Together, these observations motivate moving beyond recovering concept-like structure from trained models toward designing LLMs with explicit conceptual representations.
\end{abstract}

\section{Motivation}
\label{sec:motivation}
Large language models (LLMs) appear to encode rich concept-like structure. 
Work on neurons, latent features, and representation directions has revealed human-interpretable abstractions corresponding to semantic categories and entities \cite{shu2025survey}. Yet these abstractions emerge from distributed statistical learning rather than as explicit components of the architecture. Consequently, concepts in current LLMs are primarily \textit{discovered} after training rather than \textit{designed} into the model.

Existing interpretability methods provide valuable evidence for such structure. Sparse autoencoders (SAEs) decompose activations into human-interpretable features \citep{gurnee2026verbalizable, huben2024sparse, shu2025survey, gurnee2026verbalizable}, while probing methods identify directions associated with semantic properties and behaviors \cite{belinkov2022probing}. Yet these recovered features fall far short of human concepts. We follow the standard cognitive science usage: a concept is a mental representation that enables humans to categorize objects, events, and ideas, supports generalization and communication, and remains stable across encounters and individuals \citep{murphy2004big}. Human concepts are therefore structured, hierarchical, and compositional, expediting generalization to novel combinations. In contrast, LLM representations emerge from distributional association, with no mechanism requiring features to compose systematically or form stable, model-independent conceptual units \citep{dziri2023faith, hewitt2025position}.

This limitation is further highlighted by the instability of recovered features. Large-scale comparisons across training seeds show that only a fraction of SAE features consistently recur after retraining \citep{paulo2025sparse}, with reproducible features concentrated among a relatively small subset of functionally important representations \citep{gerasimov2026unstable}. Thus, much of the conceptual structure recovered from current LLMs appears to be an emergent property of individual models rather than an architectural commitment.

We argue that this reflects a broader limitation in how LLM research treats concepts. Rather than viewing concepts solely as objects to recover from trained models, \textbf{we propose concept-awareness as an explicit design axis that shapes how models represent, learn, and use knowledge.}

Recent work demonstrates the feasibility of this perspective, including concept-level training objectives that group semantic equivalents into shared targets \citep{iyer2026beyond, zhang2026concepts}, concept-aware fine-tuning that predicts multi-token conceptual units \citep{chen2025improving}, and architectures that introduce explicit concept representations \citep{barrault2024large, tack2025llm}. However, these approaches have emerged independently, without a unified framework for understanding where concept structure enters the LLM pipeline or how different design choices relate.

This paper organizes existing approaches into a design space defined by \emph{where} concept structure is introduced and \emph{how} concept signals are obtained. By reframing concepts as a design choice rather than a post-hoc property, we clarify existing approaches, identify unexplored directions, and outline a path toward LLMs that treat concepts as explicit computational principles.

\section{Position: Concepts as a Design Axis}
\label{sec:position}

We argue for a paradigm shift in how concept-awareness is viewed in LLMs. 
Rather than treating concepts as emergent phenomena to be recovered after training through probing, dictionary learning, or other interpretability techniques, we propose \textbf{treating concepts as deliberate design elements: computational objects that architectures can explicitly represent, manipulate, and reason over}.

This position does not prescribe a single concept-aware architecture. Instead, it elevates conceptual structure to a first-class design consideration, alongside foundational choices such as tokenization, memory mechanisms, and attention. Concept-awareness can enter the LLM pipeline at multiple stages: through objectives that encourage conceptual organization, architectures that encodes concepts, inference procedures that operate over conceptual abstractions, or post-hoc methods that recover latent structure. Together, these possibilities define a broader design space for future LLMs.

The idea of building models around human-interpretable concepts has precedents outside language modeling. Concept Bottleneck Models demonstrated that predicting explicit concepts as intermediate representations can improve interpretability and enable intervention in vision and tabular domains \citep{koh2020concept}. In contrast, LLM research has only recently begun to treat concepts as an explicit design principle at the level of core model structure \citep{sun2025concept}, despite growing interest in concept-level training objectives \citep{iyer2026beyond, zhang2026concepts}. We argue that these efforts should not be viewed as isolated techniques, but as different points within a larger landscape of concept-aware model design.

To help establish this perspective, we map where concepts \textit{have been} and \textit{can be} incorporated in the LLM pipeline. Our goal is to organize existing approaches into a common design space, identify underexplored regions, and provide a roadmap for future language models in which concepts are not merely recovered after training, but explicitly represented as part of the model's computation.

\section{Mapping Concept-Aware Interventions}
\label{sec:taxonomy}

\begin{table*}[h!]
  \centering
  \small
  \begin{tabular}{@{}p{1.55cm}|p{6.7cm}|p{6.7cm}@{}}
    \textbf{Stage} & \textbf{Internal-only} & \textbf{Externally-grounded} \\
    \midrule
    \textbf{Objective} &
    Concept-level objectives that predict shared semantic targets spanning multiple
surface forms rather than individual tokens \citep{iyer2026beyond,zhang2026concepts,liu2026next}; predicting or jointly supervising continuous concept representations rather than
relying solely on token prediction
\citep{barrault2024large,tack2025llm}&
    \emph{Sparse.} Knowledge-infused pretraining introduces external semantic supervision, but is typically framed around entities, lexical semantics, or knowledge integration rather than explicit concept-level supervision, e.g., ERNIE \cite{zhang2019ernie}, SenseBERT \cite{levine2020sensebert}, KoCo \cite{li-etal-2026-koco}, KALM \cite{rosset2020knowledge}\\
    \hline
    \textbf{Architecture} &
    Concept bottleneck models with model-derived concepts \citep{koh2020concept,sun2025concept}; architectures operating over explicit concept representations \citep{barrault2024large}; adaptive architectural routing as a precedent for designed, non-emergent computation \citep{zheng2026amor} &
    Integration of externally provided semantic structures, including knowledge graphs fused with language model layers via graph neural networks or graph-enhanced attention \citep{zhang2022greaselm,yasunaga2021qa,liu2020k,sun2020colake,zhang2022dkplm}; human-defined concept bottleneck architectures \citep{koh2020concept}\\
    \hline
    \textbf{Inference} &
    Group tokens to form concept prediction \cite{shani2023towards}; Structured recombination over parsed intermediate representations rather than token-level sampling \citep{mizrahi2025cooking}; interpretable analogy mapping over extracted representations \citep{jacob2023fame} &
    Structured inference guided by external relational representations, including graph-based reasoning traces and knowledge-augmented generation frameworks \citep{sun2024think,besta2024graph,jiang2023structgpt}\\
    \hline
    \textbf{Post-hoc} &
   Post-hoc discovery of interpretable semantic features from model representations, including sparse autoencoders \citep{shu2025survey}, monosemantic features \citep{templeton2026scaling}, emotion concepts \citep{sofroniew2026emotion}, and verbalizable directions \citep{gurnee2026verbalizable}; feature stability remains debated \citep{gerasimov2026unstable,paulo2025sparse} &
    Knowledge-graph (KG) grounded verification or explanation of an already-generated reasoning trace \cite{sansfordgrapheval}; KG-based hallucination detection and explanation \cite{dolci2026towards, niu2024mitigating, lavrinovics2025knowledge}; KG-based RAG verification \cite{he2024retrieving, jiang-etal-2025-rag} \\
  \end{tabular}
  \caption{Where and how concept structure enters into LLM pipelines. Cell contents are illustrative, not exhaustive.}
  \label{tab:grid}
\end{table*}

\emph{Where should conceptual structure enter the language modeling pipeline, and what determines the structure of that representation?} We organize concept-aware interventions along two axes.


The first captures \textbf{where conceptual structure comes from}: whether it is \emph{internally induced} from the model's own representations or \emph{externally grounded} in human-defined resources such as knowledge graphs and ontologies. This distinction matters because internally induced representations may better preserve the flexibility of neural learning, whereas externally grounded concepts lead to better interpretability and alignment.

The second axis is the \textbf{pipeline stage} at which concept structure is introduced: (i) the \emph{language modeling objective}, where concepts influence what the model is trained to predict; (ii) the \emph{core or latent architecture}, where concepts are represented explicitly within the model computation; (iii) \emph{generation-time inference}, where conceptual structure guides reasoning or generation after training; and (iv) \emph{post-hoc interpretation}, where concept-like structures are recovered from an already-trained model.

Table~\ref{tab:grid} places representative approaches within this design space. The entries should not be interpreted as claiming that each method learns human concepts in the cognitive sense; rather, they represent different strategies for introducing, inducing, or analyzing concept-like structure in LLMs.

\paragraph{A gap in the current landscape.}
Three observations emerge from this map.
First, the pipeline stages are unevenly populated. Inference is comparatively thin on both the internal and external side. This is surprising because inference is arguably where explicit concepts are most naturally useful: rather than committing to a sequence of surface-form tokens, inference procedures can operate directly over higher-level semantic units.

Second, the landscape is fragmented: concepts as explicit computational units reappears independently across pipeline stages (often without using the language of ``concepts'' at all). Concept-level objectives, structured generation procedures, and graph-augmented architectures all pursue closely related goals, yet have evolved largely in isolation. A unified perspective therefore exposes connections and opportunities that remain hidden when each line of work is viewed separately.

Third, externally grounded approaches span the \textit{entire} pipeline. Entity-infused pretraining predates graph-fused LLMs by several years, and concept-level grounding also appears in inference-time reasoning guidance and post-hoc verification. The difference is often terminology: architectural approaches use the language of \emph{concepts}, whereas similar ideas elsewhere are described as entity injection, lexical knowledge integration, reasoning guidance, or hallucination verification. Thus, the apparent dominance of architectural approaches partly reflects fragmented naming across communities rather than fundamentally different goals.

\paragraph{Beyond where concepts enter: what are concepts represented as?}
The two axes above describe \emph{where} and \emph{how} concept structure enters the LLM pipeline, but they leave open a more fundamental question: \emph{what form should a concept representation take?} Existing approaches largely instantiate concepts either as points in latent embedding spaces or as nodes in relational structures such as knowledge graphs. A third possibility is a compositional representation in which concepts are structured objects with explicit composition rules, as proposed in compositional distributional semantics \citep{coecke2010mathematical}. Such representations remain largely unexplored in modern LLMs, despite offering a direct account of how concepts combine to support systematic generalization.

Bridging these representations remains an important open direction. Recent work on sparse, interpretable correspondences between independently structured representation spaces \citep{ye2025bridged} suggests one path from embeddings to graphs, but whether similar bridges can extend to compositional representations remains unclear.

\section{Why the Map Matters}
\label{sec:why}
Separating the source of concepts from the stage at which it enters the pipeline reveals that these are independent design choices. The same conceptual structure can be introduced at different pipeline stages, while the same stage can operate over either internally induced or externally grounded concepts. This distinction lets researchers deliberately choose an entry point. Objective-level interventions are comparatively inexpensive and have shown gains in human alignment and robustness \citep{iyer2026beyond, zhang2026concepts}, but leave the resulting architecture opaque. Core architectural interventions make concepts directly addressable and editable, but are more invasive. Generation-time interventions are comparatively non-invasive and naturally fit non-autoregressive and parallel-refinement paradigms, but incur latency \citep{mizrahi2025cooking, shani2023towards}. Post-hoc interpretation requires no architectural commitment, but, as seed-instability findings show, offers the weakest guarantee that what is found is what is actually there \citep{gerasimov2026unstable, paulo2025sparse}.

We note that some architectural work gestures at higher-granularity representations for efficiency reasons alone, without any concept-level claim. For example, hourglass shortens token sequences into coarser units to reduce attention cost \citep{nawrot-etal-2022-hierarchical}, illustrating that ``operating above the token level'' and ``operating over human concepts'' are separate properties that are easy to conflate.

\section{Open Challenges}
\label{sec:challenges}

Our design space reveals four challenges that cut across approaches, regardless of where concept structure enters the pipeline or whether it is internally discovered or externally grounded.

\paragraph{Compositionality.}
Making concepts explicit is not enough to make them useful. Every cell in Table~\ref{tab:grid} can, in principle, produce addressable conceptual units; however, \textbf{explicitness alone does not provide the compositional structure underlying human-like generalization}. Human concepts support novel combinations through systematic compositional principles \citep{murphy2004big}. A concept bottleneck, graph node, or self-supervised concept objective may expose meaningful units, but none inherently guarantees that they interact through reusable rules rather than learned associations. Transformer LLMs illustrate this gap: despite strong performance on many compositional tasks, they often rely on memorized patterns, with performance degrading as compositional complexity increases \citep{dziri2023faith}. As discussed in \S~\ref{sec:taxonomy}, a genuinely compositional representation remains unrealized in LLM architectures \citep{coecke2010mathematical}; whether any pipeline stage or structural source in our map can support it remains open.

\paragraph{Found versus designed.}
Introducing concepts into an architecture does not resolve the fundamental question of \textbf{what it means for a model to represent a concept}. We still need criteria that distinguish a concept that a model genuinely encodes from a feature that is merely correlated with a probe, objective, or graph node used to elicit it. Recent work on feature stability provides one possible methodological direction: testing whether representations persist under controlled perturbations such as changes in random initialization \citep{gerasimov2026unstable}. However, such analyses have focused on post-hoc interpretability methods. Extending analogous evaluation principles to concept bottlenecks, concept-aware objectives, or graph-integrated architectures remains an open challenge.

\paragraph{Whose concepts?}
Concept-awareness requires a decision about where conceptual boundaries come from. Externally-grounded approaches impose structure from human-designed ontologies or other sources of prior knowledge. Internally-derived approaches allow conceptual categories to emerge from the model's own experience and learning dynamics. Neither choice is universally preferable: \textbf{externally imposed concepts may improve interpretability and alignment while constraining what the model can discover, whereas internally derived concepts may capture model-specific regularities while potentially diverging from human conceptual organization}. The appropriate balance depends on the intended use of the model, yet this design choice remains largely implicit.

While externally imposed concept structure can improve interpretability and alignment, it also introduces inductive biases that may be inappropriate for complex or weakly defined semantic domains. If conceptual abstractions are fixed prematurely, models may collapse distinctions that would otherwise emerge through experience, potentially limiting representational flexibility. Whether scaling, continued post-training, or adaptive concept learning mitigates this tradeoff remains an open question.

\paragraph{No shared benchmark across cells.}
Because approaches at different pipeline stages have evolved largely independently, the field lacks a common framework for comparing concept-based designs. No benchmark currently enables pre-, mid-, and post-training interventions, architectural modifications, and inference-time methods to be evaluated on shared dimensions such as conceptual fidelity, stability, compositionality, and downstream utility. \textbf{Establishing such a benchmark is essential for turning concept-awareness from a collection of promising ideas into a mature design paradigm.}

\section*{Limitations}
This position paper offers a map of the design space rather than an exhaustive survey or a new empirical method; the grid in Table~\ref{tab:grid} is illustrative and almost certainly incomplete, particularly for the sparser cells. We also do not adjudicate which pipeline stage or structural source is best for a given application, since that judgment depends on constraints (compute, need for post-hoc auditability, tolerance for architectural change) that vary by setting.

\bibliography{custom}

\end{document}